\documentclass{article}

     \usepackage[final]{tackling_climate_workshop_style}

\usepackage[utf8]{inputenc} 
\usepackage[T1]{fontenc}    
\usepackage{url}            
\usepackage{booktabs}       
\usepackage{amsfonts}       
\usepackage{nicefrac}       
\usepackage{microtype}      

\usepackage{graphicx}
\usepackage{amsmath}
\usepackage{amssymb}
\usepackage{parskip}
\usepackage{booktabs}
\usepackage{multirow}
\usepackage{caption} 
\usepackage{subfigure}
\usepackage{tikz,graphics,color,float,epsf,caption}
\usepackage[backref=page]{hyperref}
\usepackage{enumitem}

\title{Weakly-semi-supervised object detection\\in remotely sensed imagery}

\author{Ji Hun Wang\thanks{Equal contribution.}, Jeremy Irvin\footnotemark[1], Beri Kohen Behar, Ha Tran,\\
\textbf{Raghav Samavedam, Quentin Hsu, Andrew Y. Ng}\\
Stanford University\\
{\tt\small \{jihunwang,jirvin16,bkohen,hahntrn,raghavsa,qhsu,ang\}@cs.stanford.edu}
}

\begin{document}

\maketitle

\begin{abstract}
Deep learning for detecting objects in remotely sensed imagery can enable new technologies for important applications including mitigating climate change. However, these models often require large datasets labeled with bounding box annotations which are expensive to curate, prohibiting the development of models for new tasks and geographies. To address this challenge, we develop weakly-semi-supervised object detection (WSSOD) models on remotely sensed imagery which can leverage a small amount of bounding boxes together with a large amount of point labels that are easy to acquire at scale in geospatial data. We train WSSOD models which use large amounts of point-labeled images with varying fractions of bounding box labeled images in FAIR1M and a wind turbine detection dataset, and demonstrate that they substantially outperform fully supervised models trained with the same amount of bounding box labeled images on both datasets. Furthermore, we find that the WSSOD models trained with 2-10x fewer bounding box labeled images can perform similarly to or outperform fully supervised models trained on the full set of bounding-box labeled images.
We believe that the approach can be extended to other remote sensing tasks to reduce reliance on bounding box labels and increase development of models for impactful applications.
\end{abstract}

\section{Introduction}

Machine learning for detecting objects in remotely sensed imagery has demonstrated success across diverse tasks \cite{cheng2016survey, hoeser2020object} including those for impactful applications like sustainable development \cite{burke2021using} and tackling climate change \cite{rolnick2022tackling}. However, many such approaches use object detection models that require large amounts of bounding box labels which can be costly and time-consuming to obtain, especially for remote sensing tasks that require specialized expertise to label. This dependency on bounding box labels substantially limits the development of models for new tasks and new geographies \cite{zhu2017deep}.

Semi- and weakly-supervised object detection approaches have been developed to alleviate the need for large amounts of bounding box labels. Semi-supervised object detection makes use of a small number of bounding box annotations and a large amount of unlabeled images \cite{sohn2020simple, zhou2021instant}. Semi-supervised methods have been applied to remotely sensed imagery with mixed levels of success \cite{hong2020x, wang2020semi}.
Many semi-supervised approaches adopt a student-teacher framework: the teacher is trained on the limited supervised data and then run on unsupervised data to produce pseudo-labels, which are then used to train the student model. 
Alternatively, weakly-supervised object detection uses weaker forms of supervision, commonly including image-level annotations with no bounding boxes \cite{bilen2016weakly, li2019weakly, tang2018pcl, wan2019c, yan2017weakly} and points \cite{ren2020ufo}. Many prior works have investigated weakly-supervised learning on remotely sensed imagery using image-level labels \cite{wang2020weakly, han2014object, mohammadi2020deepwind} and noisy labels \cite{zhang2020weakly, cheng2020weakly}. However, to our knowledge, no works have explored weakly-semi-supervised methods for remote sensing tasks.

Geographic point locations are highly prevalent in existing public datasets and have the potential to serve as `free' labels for training object detection models. Point locations are commonly collected as part of geospatial datasets and constitute a large portion of the OpenStreetMap (OSM) database \cite{mooney2017review, bennett2010openstreetmap}. These point locations can be leveraged as weak labels to develop models with a minimal number of additional bounding box labels, but the vast majority of object detection architectures have not been designed to use point labels as supervision.

Recently, approaches have been emerging to leverage a large amount of point annotations together with a limited amount of bounding box labels, referred to as weakly-semi-supervised object detection with points (WSSOD-P) \cite{bearman2016s, chen2021points, zhang2022group, yan2017weakly}. A general pipeline for WSSOD-P used in recent works \cite{chen2021points, zhang2022group} consists of three parts. First, a teacher model is trained using a limited set of strongly labeled data to input an image with a set of point annotations and produce a corresponding set of bounding boxes. Second, the teacher model is applied to a large dataset of images with point labels to generate pseudo-bounding box annotations on the images. Third, a supervised object detection model is trained on both the strongly labeled images and the pseudo-labeled images. 

In this work, we use Group R-CNN, the state-of-the art in WSSOD-P on natural images at the time of experimentation \cite{zhang2022group}. Group R-CNN follows the three stage pipeline for WSSOD-P described above but proposes a few key changes to the point-to-box regressor, namely producing multiple proposals for each point to improve recall, instance-level proposal assignment to improve instance differentiation and therefore precision, and finally instance-aware representation learning to handle convergence issues with instance assignments \cite{zhang2022group}. Group R-CNN achieves high performance on MS-COCO, but its performance on remote sensing tasks has not been investigated prior to this work.

\begin{figure*}
\centering

\subfigure[Teacher model training and inference procedure.]{\label{fig:location-a}
\includegraphics[width=0.50\textwidth]{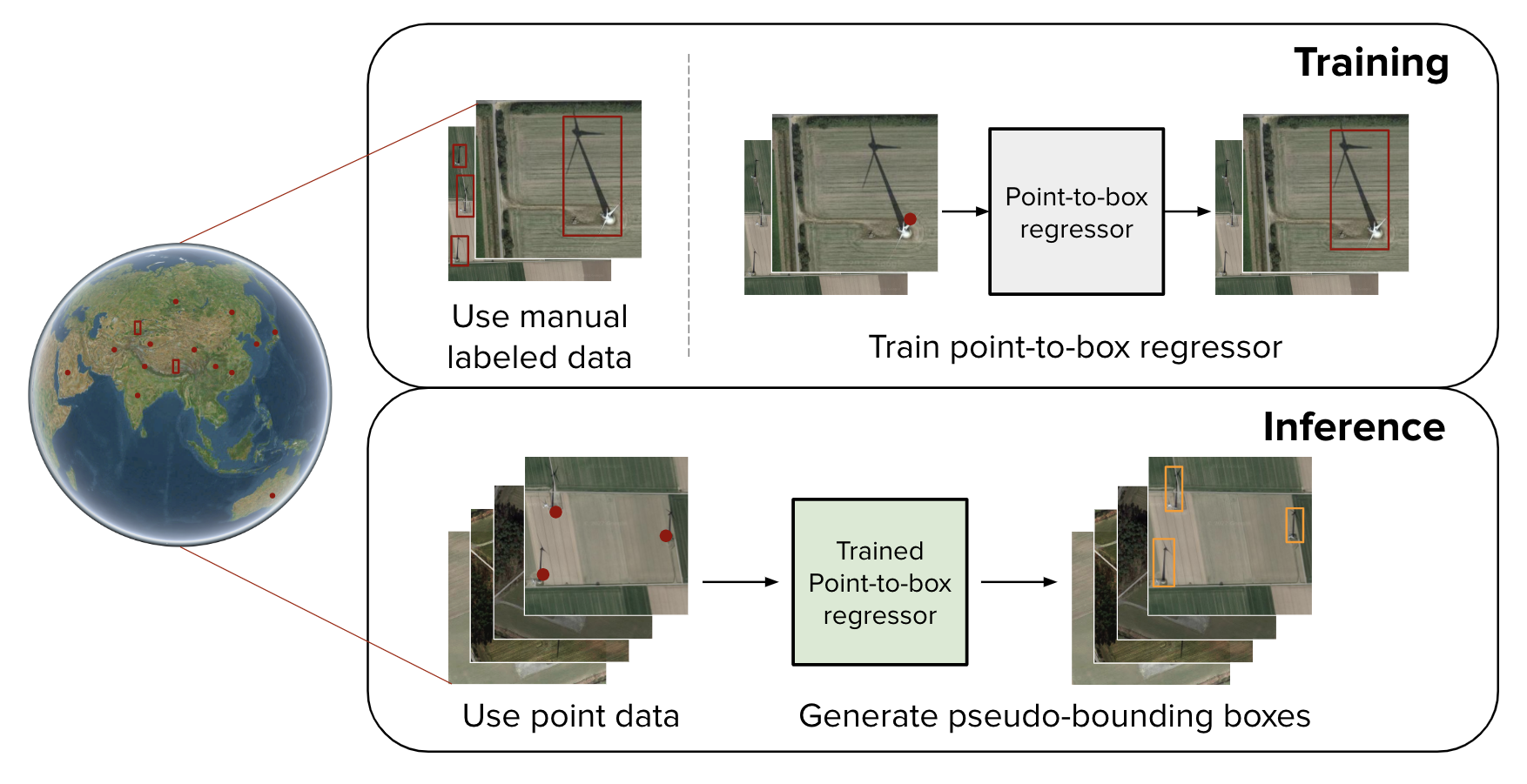}}
\subfigure[Student model training and inference procedure.]{\label{fig:location-b}
\includegraphics[width=0.45\textwidth]{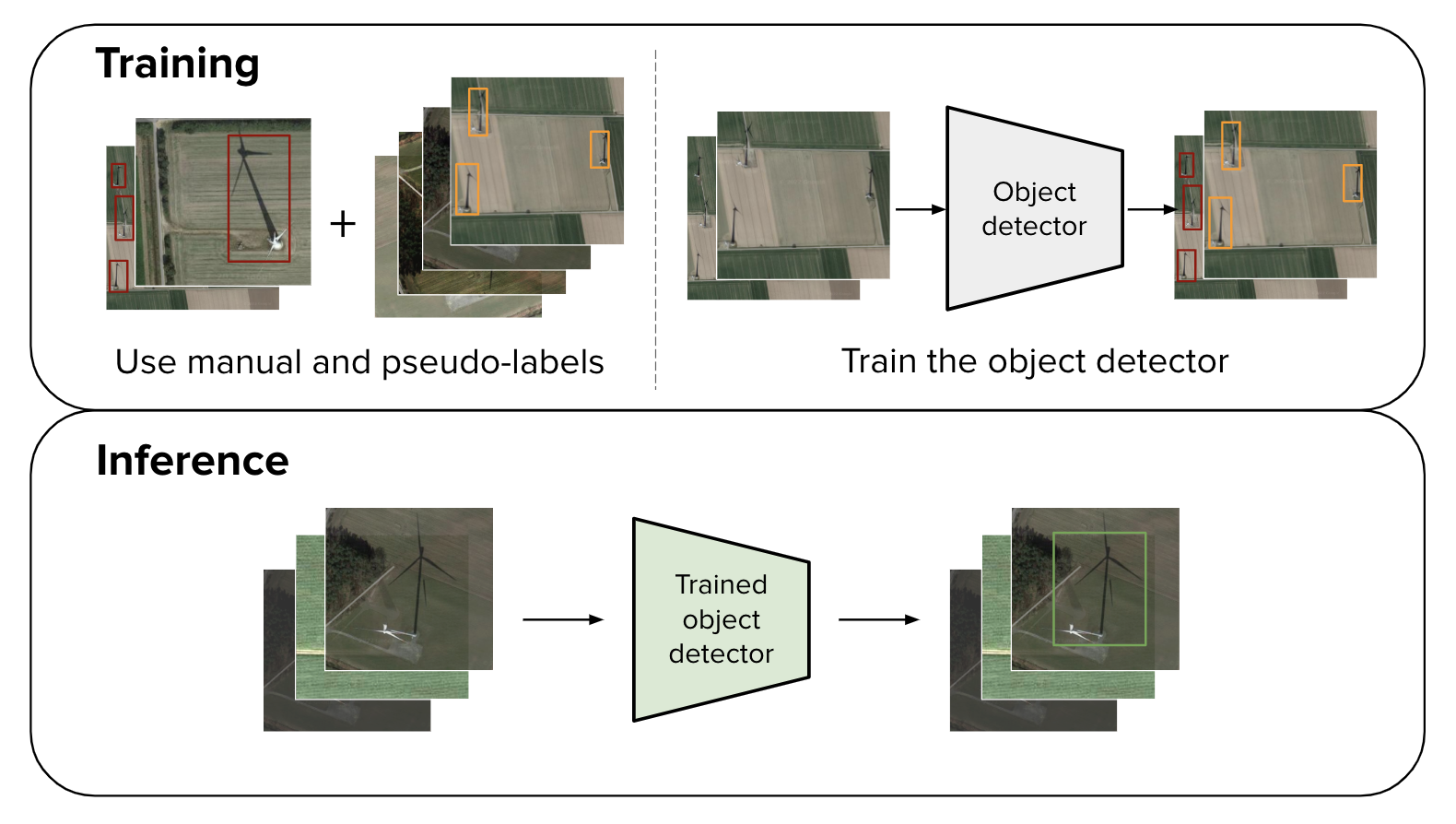}}
\caption{Overview of the weakly-semi-supervised object detection training and inference procedures of the teacher and student models.}
\label{fig:overview}
\end{figure*}

Our contributions are as follows:
\begin{enumerate}[itemsep=-2mm]
\item We develop Group R-CNN models for WSSOD-P (Figure~\ref{fig:overview}) on two remote sensing datasets.
\item We validate the approaches using different amounts of bounding box labeled images, and demonstrate the inclusion of point labels substantially improves performance when compared using the same amount of bounding box labels across all label amounts, with some models outperforming fully supervised models trained with 2-10x more bounding box labeled data.
\item We additionally explore the effect of geographic distribution shift on the performance of the approach for wind turbine detection, and find that although the approach provides benefits compared to fully supervised training, it is still susceptible to the distribution shift.
\end{enumerate}

 We believe that the approach can be extended to many different remote sensing tasks to substantially reduce reliance on strongly labeled datasets.

\section{Methods}





\subsection{Datasets}
We use two remote sensing datasets to develop our approaches: an internally developed dataset of satellite imagery capturing wind turbines and the publicly available FAIR1M dataset \cite{sun2022fair1m}. We select the first dataset as it has direct applications to mitigating climate change \cite{zhou2019deepwind} and serves as a simple benchmark with the task of detecting a single type of object with intra-class homogeneity, while the second dataset is a more complex detection task with multiple classes with intra-class heterogeneity. We follow the provided training, validation, and test splits for FAIR1M but experiment with two different types of splits on the wind turbine dataset, one split randomly by wind farms (in-country) and one split by country to investigate the effect of geographic distribution shift (out-country). More detail about dataset processing is provided in the Appendix.

\subsection{Weakly-Semi-Supervised Model}
We adopt Group R-CNN as the weakly-semi-supervised model in our experiments \cite{zhang2022group}. The architecture consists of two models: a teacher model which inputs an image with point labels and outputs pseudo-bounding box labels for each point, and a student model which is trained on the pseudo-labeled images together with the manually labeled images. The training and inference procedures are shown in Figure~\ref{fig:overview}, and training procedure details for all models are provided in the Appendix.

\subsubsection{Teacher Model}
We use the same settings for the teacher model architecture used in the original Group R-CNN work, including a ResNet-50 backbone \cite{he2016deep} and RetinaNet as the region proposal detector \cite{lin2017focal} with an IoU threshold of 0.7. Furthermore, we use a category embedding with 1 class (functionally equivalent to no category embedding) for the wind turbine datasets and 5 classes for FAIR1M.

\subsubsection{Student Model}
We use the same architectural settings for the student model used in the original Group R-CNN, including a ResNet-50 backbone, Feature Pyramid Network (FPN) neck \cite{lin2017feature}, and a Fully Convolutional One-Stage (FCOS) object detection head, where the class, bounding box, and centerness loss are calculated using focal loss, IoU loss, and cross-entropy loss, respectively. For the wind turbine datasets which are single class detection tasks, we modify the architecture to remove the classification head. For FAIR1M, we replace the 80 category classification head with one that produces 5 classification scores specific to the FAIR1M categories.

\subsection{Experiments}

\noindent
\textbf{Fraction of Images with Bounding Box Annotations.} \, To investigate the impact of leveraging weakly labeled data with varying amounts of bounding boxes labeled in the model, we experiment with varying fractions of the images with bounding boxes and use the remaining images with point annotations as the weakly labeled data. For both the in-country and out-country wind turbine datasets, we use approximately 1\%, 5\%, and 10\% of the dataset as bounding box labeled images with 99\%, 95\%, and 90\% of the dataset as point labeled images, respectively. These fractions correspond to 515, 2433, and 5530 images respectively for the in-country dataset and 418, 2094, and 4187 respectively for the out-country dataset. For FAIR1M, we try to approximately match the wind dataset counts and use 10\% (1104 images) and 50\% (5520 images) of the dataset as bounding box labeled images with 90\% and 50\% of the dataset as point labeled images respectively. We do not use less than 10\% on FAIR1M as it leads to highly unstable results, likely due to the lack of examples for some classes.

\noindent
\textbf{Evaluation Metrics.} \, We measure the mean average precision (mAP) of bounding box predictions computed over three different intersection-over-union (IoU) thresholds, namely 0.25, 0.5, and 0.75. For the wind turbine datasets, we refer to the mAP as AP since the task is single class.
\section{Results}
We report the performance of the Group R-CNN student model and the performance of the fully supervised object detection model across varying proportions of bounding box labeled images below. The teacher model results are reported in the Appendix.


\begin{figure*}
\centering
\subfigure[In-country Wind Turbine]{\label{fig:student-a}\includegraphics[width=0.32\textwidth]{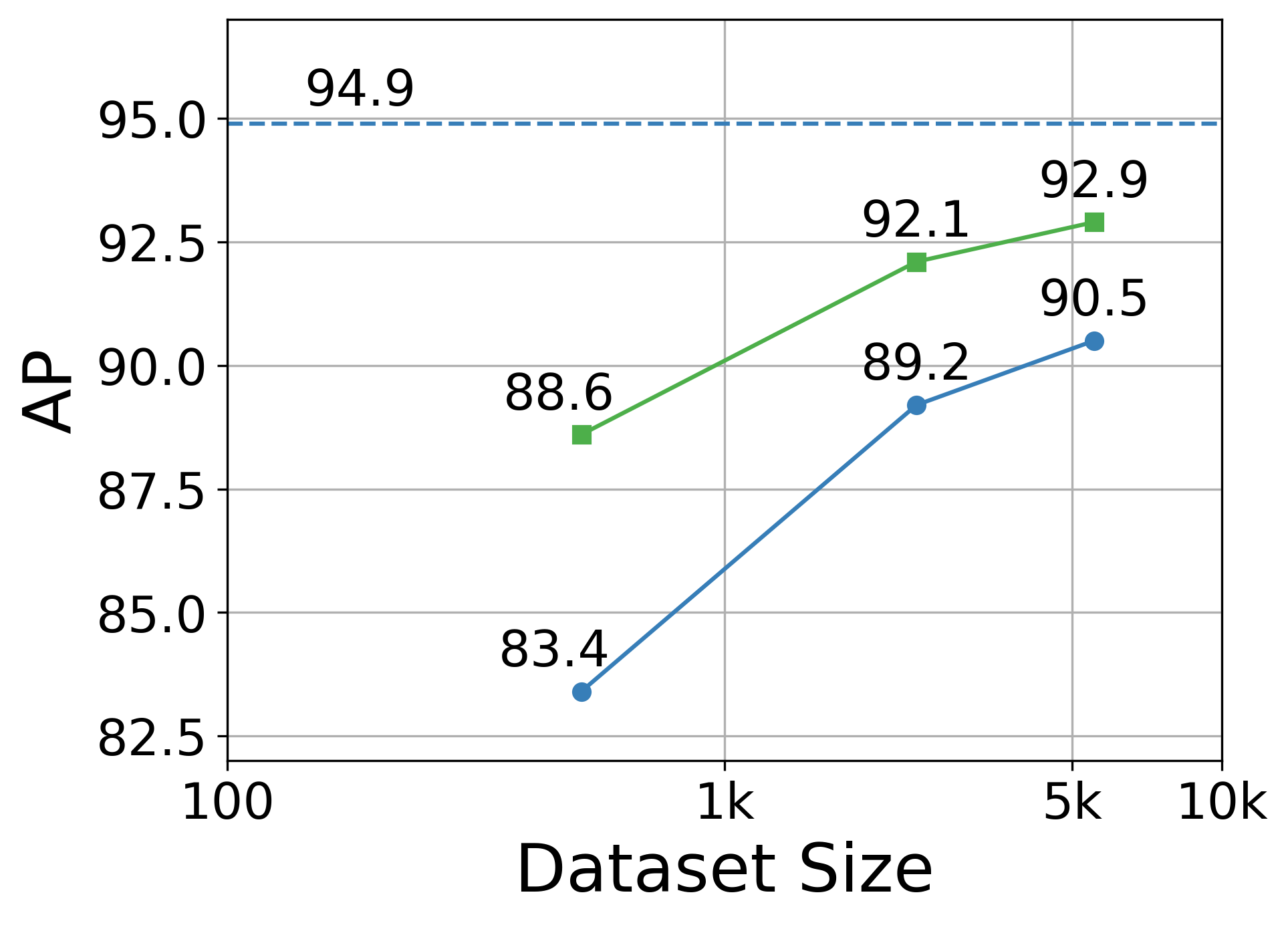}}
\subfigure[Out-country Wind Turbine]{\label{fig:student-b}\includegraphics[width=0.32\textwidth]{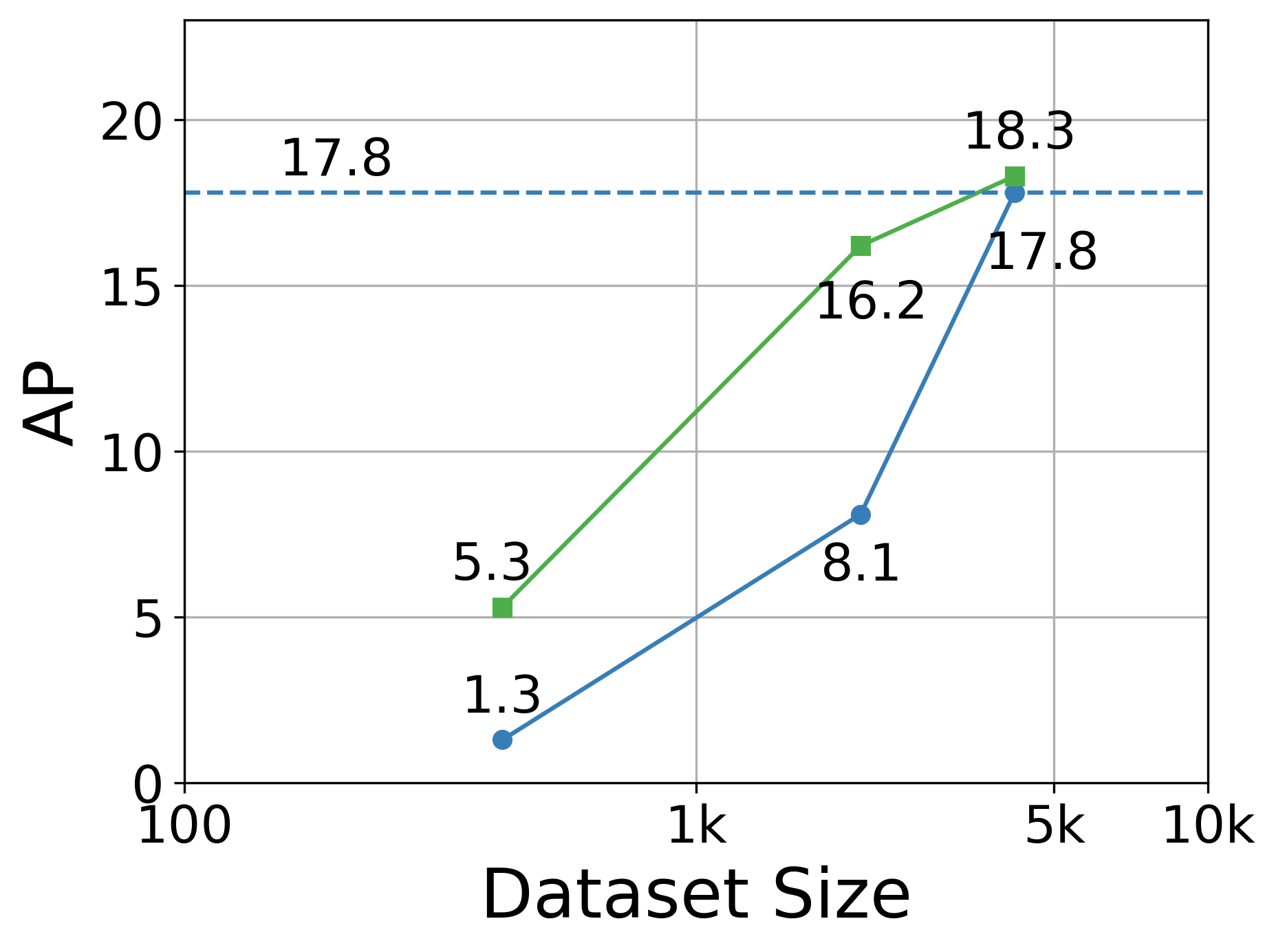}}
\subfigure[FAIR1M]{\label{fig:student-c}\includegraphics[width=0.32\textwidth]{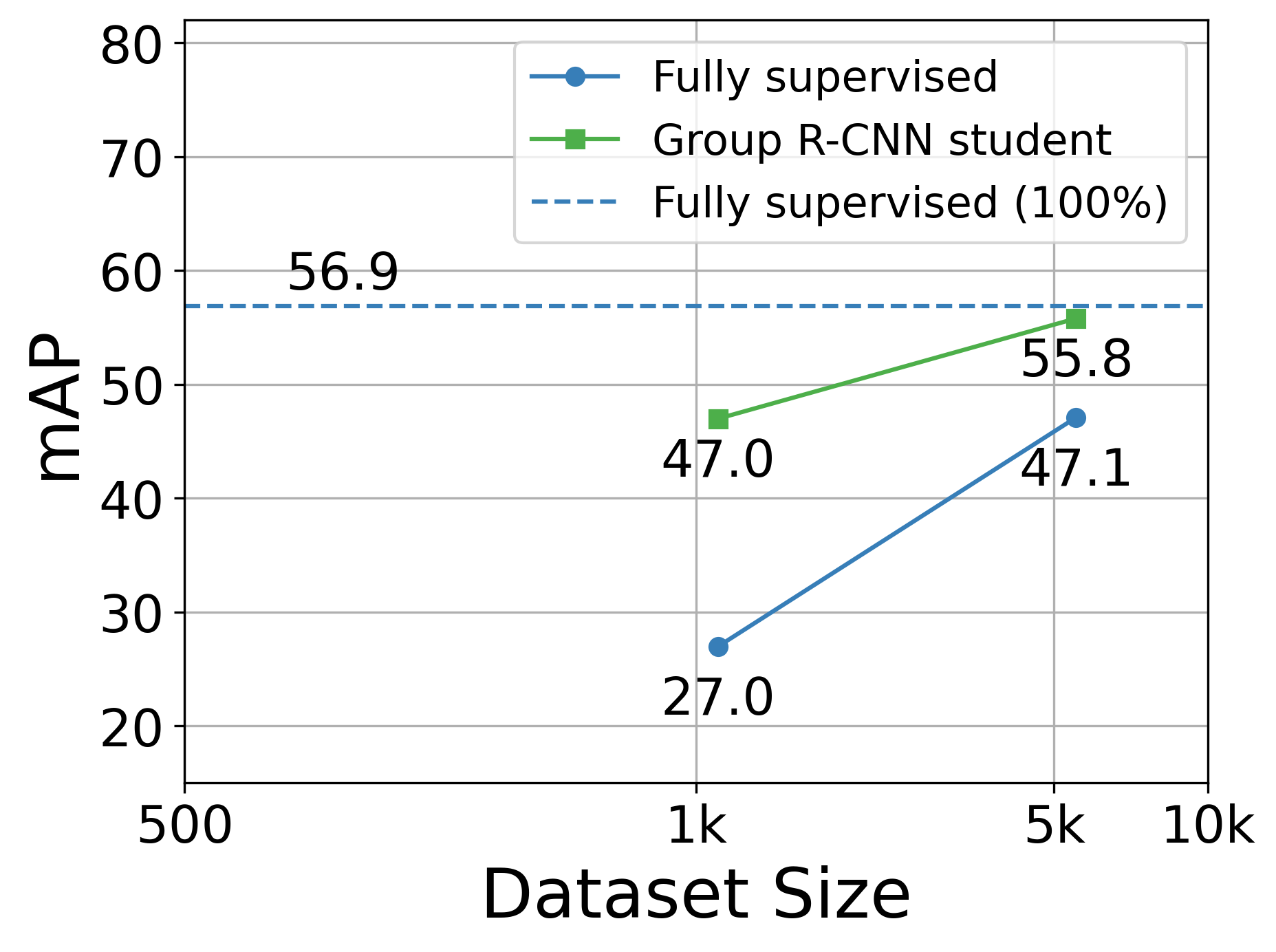}}
\caption{Comparison of Group R-CNN student model in comparison to fully supervised counterparts on the test sets of each dataset. mAP is evaluated using a 0.5 IoU threshold.
}
\label{fig:student}
\end{figure*}

\subsection{In-country Wind Turbine}

The student model outperforms the fully supervised model across all fractions of bounding box labeled images (Figure~\ref{fig:student-a}). The performance improvement is highest when using 1\% of the bounding box labeled data ($+$5.2 AP) compared to 5\% ($+$2.9 AP) and 10\% ($+$2.4 AP). The same trends hold for AP at the other IoU thresholds (Figure~\ref{fig:other_student}).

The student model trained with fewer bounding box labeled data often performs similarly to or outperforms the fully supervised model trained with more bounding box labeled data. The 1\% student model slightly underperforms the 5\% fully supervised model ($-$0.6 AP) and the 5\% student model outperforms the 10\% fully supervised model ($+$1.6 AP). Both the 5\% student model and the 10\% student model moderately underperform the fully supervised model trained with 100\% of the bounding box labeled data ($-$2.8 AP and $-$2.0 AP, respectively).




\subsection{Out-country Wind Turbine}

The student model also consistently outperforms its fully supervised counterparts on the out-country turbine dataset (Figure ~\ref{fig:student-b}). The performance improvement is highest when using 5\% of the bounding box labeled data ($+$8.1 AP) compared to 1\% ($+$4.0 AP) and 10\% ($+$0.5 AP). Notably, the student performance on the out-country set is much lower than the student performance on the in-country set, which is likely a consequence of geographic distribution shift.

However, even under this distribution shift, the student model trained with 5\% and 10\% of the manual labels achieves similar or better performance than the fully supervised model trained with all the bounding box labels: the 5\% student model moderately underperforms the 100\% fully supervised model ($-$1.6 AP) and the 10\% student model slightly outperforms it ($+$0.5 AP).




\subsection{FAIR1M}
The student outperforms the fully supervised model across both data fractions on FAIR1M
(Figure~\ref{fig:student-c}). It massively outperforms the fully supervised model with 10\% of the bounding box labeled images ($+$20.0 mAP) and outperforms at 50\% by $+$8.7 mAP. Notably, the 50\% student model only underperforms the 100\% fully supervised model by $-$1.1 mAP. Furthermore, the performance of the 10\% student model almost matches that of the 50\% fully supervised model ($-$0.1 mAP). 

\section{Discussion}
The results demonstrate the benefit of using weakly-semi-supervised object detection to reduce reliance on manually labeled bounding box data.
The student model trained with pseudo-labels outperforms the fully supervised counterpart on all tested data fractions on three datasets capturing inter- and intra-class heterogeneity as well as geographic distribution shift. It is worth noting that the out-country wind turbine performance is substantially lower than the in-country performance for both the teacher and student models, suggesting that geographic distribution shift indeed plagues these models. It is possible that this approach can be combined with strategies to address the distribution shift which have been widely explored in the field of computer vision and in remote sensing \cite{liu2022deep,tuia2016domain}.

Our approach can be extended to other remote sensing tasks to reduce reliance on strongly labeled data, which we hope will help enable the development of more remote sensing models for impactful applications. Importantly, remote sensing has played and will continue to play a major role for mitigating and adapting to climate change \cite{zhao2023role}, and AI is a key tool to enable and accelerate methods for automatically extracting insights from the immense amounts of remote sensing data \cite{zhang2022artificial}. We believe approaches like ours to reduce the need for labels will continue to help advance these technologies.



\newpage
{\small
\bibliographystyle{ieeetr}
\bibliography{tackling_climate_workshop}
}

\appendix
\newpage
\section*{Appendix}
\subsection*{Dataset Processing}
\noindent
\textbf{Wind Turbine Datasets.} \, The wind turbine dataset consists of 200,000 Airbus SPOT RGB images capturing 85,000 wind turbines and 110,000 images without any turbines spanning 40 countries. We curate the dataset by obtaining globally distributed wind turbine point locations from OpenStreetMap and constructing 416 $\times$ 416 Airbus SPOT RGB images capturing the wind turbines as well as randomly sampled locations using the Descartes Labs platform \cite{beneke2017platform, haklay2008openstreetmap, bennett2010openstreetmap}. We construct the images to share no spatial overlap with any other image in the dataset. We then upload the positive images to a crowdsourcing platform to obtain bounding box labels around the turbines. Out of these images, we only retain the ones where at least one bounding box label contains an OpenStreetMap coordinate; point annotations in these retained images which were not labeled with bounding boxes are dropped from the dataset. A single image may contain multiple wind turbines and each wind turbine is associated with a bounding box and the original OpenStreetMap point annotation, which mostly occurs at the base of the wind turbine.

We split the wind turbine dataset into a training set, a validation set, and a test set in two different ways. Before splitting the dataset, we first cluster the wind turbines into groups (which serve as a proxy for wind farms) using DBSCAN with Haversine distance as the distance metric and a group of 3 turbines as the minimum cluster size. We then split the wind turbine clusters into training, validation, and test sets in order to ensure wind turbines from the same farm occur in the same split, as turbines and land use within a farm tend to share very similar features. For the first splitting strategy, we randomly split the clusters into training, validation, and test sets and refer to this dataset as the ``in-country wind turbine dataset.''

In a practical setting, it is common to only have bounding boxes in one set of regions and points everywhere else, which may lead to point-to-box regressor generalization issues. To test the effect of this ``geographic distribution shift'' due to changes in turbine appearance and landscapes in different countries, we adopt a second splitting strategy where we split the clusters by country such that wind turbines within one country appear in the same split. As China, United States, and Spain combined account for roughly 85\% of the images in the dataset, we sample the training set to consist of all turbines in the western half of the United States, the validation set to consist of all turbines in the eastern half of the United States, and the test set to consist of all wind turbines located in the 37 countries other than China, United States, and Spain. To report the teacher model's performance, we instead use a separate set that consists of images of wind turbines located in China and Spain. We refer to these splits as the ``out-country wind turbine dataset.''

The in-country dataset contains 200,000 images in the training set (56,000 positive images), 54,000 images in the validation set (19,000 positive images), and 59,000 images in the test set (21,000 positive images). The out-country dataset contains 150,000 images in the training set (60,000 positive images), 50,000 images in the validation set (13,000 positive images), and 50,000 images in the test set (15,000 positive images).

\vspace{5pt}
\noindent
\textbf{FAIR1M.} \quad FAIR1M is a publicly available object detection dataset containing high resolution satellite imagery labeled with bounding boxes of objects from a variety of categories. It contains more than 15,000 fully annotated images and more than 1 million instances, where each object in an image belongs to one of five main categories (ship, vehicle, airplane, sports court, and road) and 37 subcategories \cite{sun2022fair1m}. The number of objects embedded within an image varies from 1 to 596. The sizes of images also vary: 10 images are of size 7000 $\times$ 7000, while the majority of images have width and height of 1000 or smaller.  Due to constraints with computational resources, we drop images if they have a resolution higher than 2000 $\times$ 2000 or more than 100 object annotations. This results in a total of 11,804 images in the dataset. We convert the oriented bounding boxes to axis-aligned bounding boxes using the minimum bounding rectangle around the original boxes. The point labels used in our experiments are the centers of the corresponding axis-aligned bounding boxes. We follow the same splits defined by the data providers.

\subsection*{Model Training Procedure}
For all experiments, we use an SGD optimizer with a momentum of 0.9, weight decay of 0.0001, and a batch size of 2. The initial learning rate was tuned for each model and dataset by experimenting with learning rates 0.001, 0.002, ..., 0.01. We use the same learning rate scheduler as \cite{zhang2022group}, where learning rate decay occurs at the 30th and 40th epoch. We use random data augmentations including resize, random flip, crop, and normalization. All models were trained using PyTorch on four A4000 GPUs.

\subsection*{Teacher Model Results}

\begin{table*}[t!]
\centering
\begin{tabular}{@{}ccc ccc}
\toprule
    & &  &\multicolumn{3}{c}{mAP} \\
    & & Fraction & IoU=0.25 & IoU=0.5 & IoU=0.75 \\ \midrule
    & \multirow{3}{*}{In-country Wind Turbine} 
                                & 1\% & 96.7 & 89.5 & 50.7 \\
                              &  & 5\% & 97.9 & 93.4 & 69.5 \\
                              &  & 10\% & 97.9 & 94.4 & 72.4 \\ \midrule
    & \multirow{3}{*}{Out-country Wind Turbine} 
                                 & 1\% & 31.0 & 3.7 & 0.1 \\
                              &   & 5\% & 47.7 & 18.8 & 4.4 \\
                              &   & 10\% & 55.0 & 23.1 & 6.6 \\ \midrule
    & \multirow{2}{*}{FAIR1M} & 10\% & 86.8 & 72.6 & 36.1 \\
                              &   & 50\% & 93.2 & 84.8 & 55.1 \\\bottomrule

\end{tabular}
\caption{Results of the teacher model on the test set of the three remote sensing datasets in terms of average precision. Performance is reported at multiple IoU thresholds to measure the preciseness of the teacher model's bounding box predictions across the label fractions.}
\label{table:teacher}
\end{table*}

\subsubsection*{In-country Wind Turbine}
\begin{figure*}[!htb]
  \centering
  
  \begin{tabular}{ccc}{\includegraphics[width=0.30\textwidth]{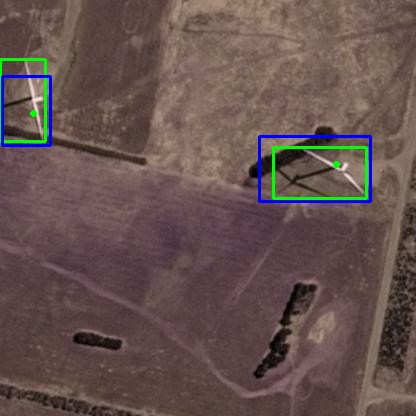}} &{\includegraphics[width=0.30\textwidth]{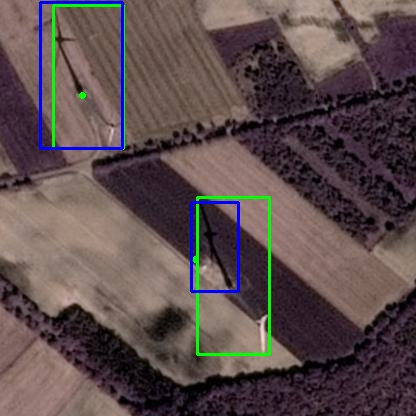}} &{\includegraphics[width=0.30\textwidth]{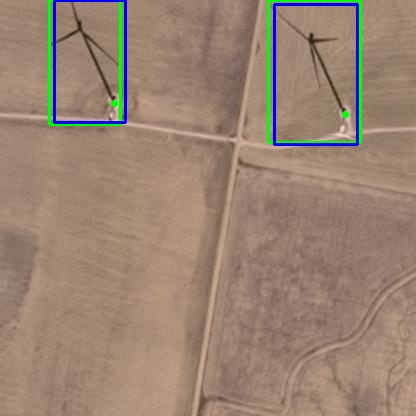}} \\
    
    \includegraphics[width=0.30\textwidth]{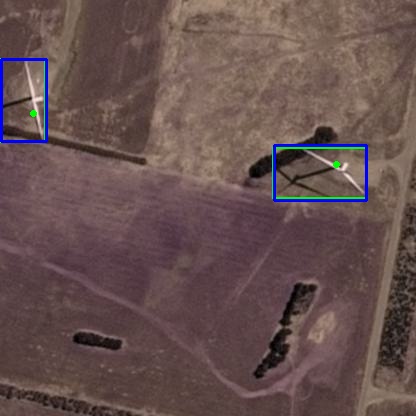} &
    \includegraphics[width=0.30\textwidth]{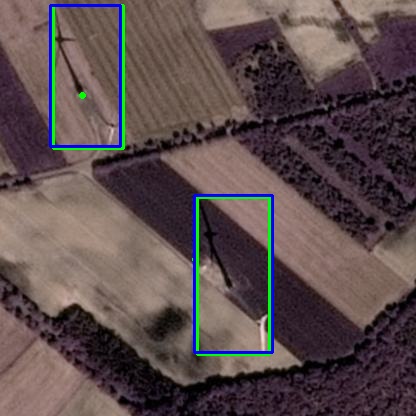} &
    \includegraphics[width=0.30\textwidth]{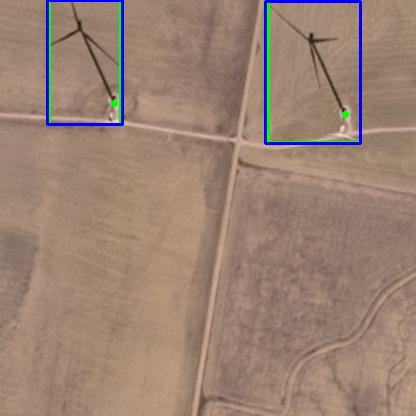} \\
  \end{tabular}
  
  \caption{Teacher predictions of the in-country wind turbine model with 1\% (first row) and 10\% (second row) data. Predicted bounding boxes are visualized in blue and ground truth boxes are visualized in green.}
  \label{fig:example}
\end{figure*}

The performance of the teacher model on the in-country wind turbine test set is $>$96 AP across all tested fractions at an IoU of 0.25 (Table~\ref{table:teacher}). At an IoU of 0.5, teacher performance remains high with all methods achieving close to or above 90 AP. There is a substantial drop in performance across all data fractions when using an IoU threshold of 0.75, suggesting difficulties with producing precise bounding boxes, especially when using 1\% of the data (see Figure~\ref{fig:example} for examples).

Improvement differences between fractions are more apparent at higher IoU thresholds. At an IoU threshold of 0.25, using 5\% of the bounding box data leads to a small improvement ($+$1.2 AP) compared to 1\% of the data, and there is no improvement from 5\% to 10\%. At an IoU of 0.5, we observe a moderate performance improvement from 1\% to 5\% ($+$3.9 AP) and a small improvement from 5\% to 10\% ($+$1.0 AP). The improvement from 1\% to 5\% at an IoU of 0.75 is especially large ($+$18.8 AP) and moderate from 5\% to 10\% ($+$2.9 AP).

To help interpret the differences in performance between teacher models trained with varying fractions of data, we visualize the predictions of the teacher models trained with 1\% and 10\% of the data (Figure ~\ref{fig:example}). The 1\% model often fails to capture the full extent of the wind turbine and shadow, whereas the 10\% model consistently captures their full extents.

\subsubsection*{Out-country Wind Turbine}
The teacher model performance on the out-country wind turbine test set is substantially lower than that of the in-country wind turbine test set (Table~\ref{table:teacher}), demonstrating the impact of the geographic distribution shift. We see substantial performance improvements from including more data across all IoU thresholds. Specifically, at an IoU threshold of 0.5, we see a 15.1 improvement in AP by increasing the amount of data from 1\% to 5\% and a 4.3 improvement in AP by increasing it from 5\% to 10\%. We observe similar trends at the other IoU thresholds. 



\subsubsection*{FAIR1M}

The performance of the teacher model on the FAIR1M dataset $>$85 mAP across both tested data fractions at an IoU threshold of 0.25 (Table~\ref{table:teacher}). Performance of the teacher model with both data fractions is $>$70 mAP at an IoU threshold of 0.5 and drops substantially when evaluating using an IoU threshold of 0.75. We observe an improvement of 6.4, 12.2, and 19 AP from using 10\% to 50\% of the bounding box data at IoU thresholds of 0.25, 0.5, and 0.75, respectively.

\begin{figure*}
\centering
\subfigure[In-country Turbine, IoU=0.25]{\label{fig:other25_student-a}\includegraphics[width=0.32\textwidth]{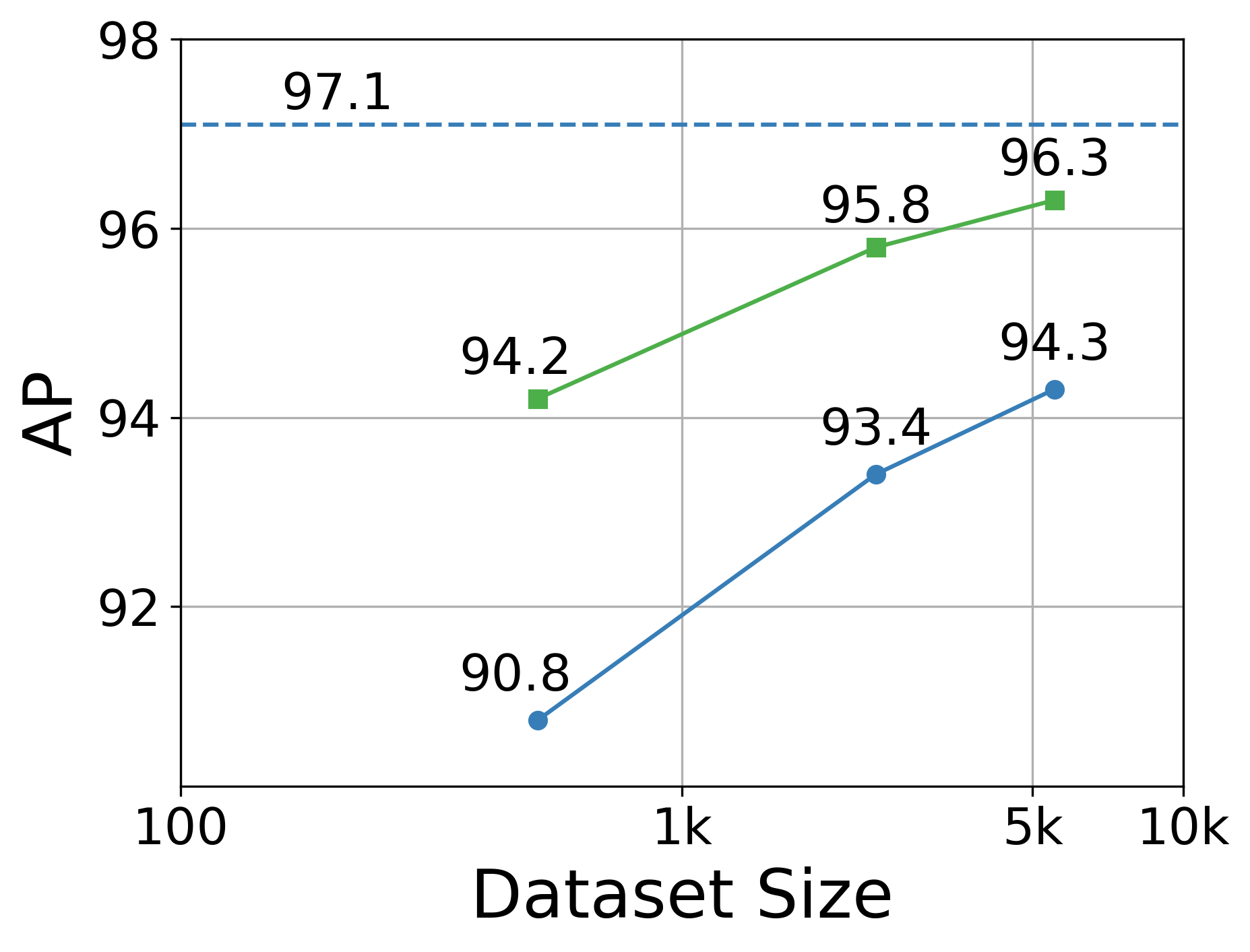}}
\subfigure[Out-country Turbine, IoU=0.25]{\label{fig:other25_student-b}\includegraphics[width=0.32\textwidth]{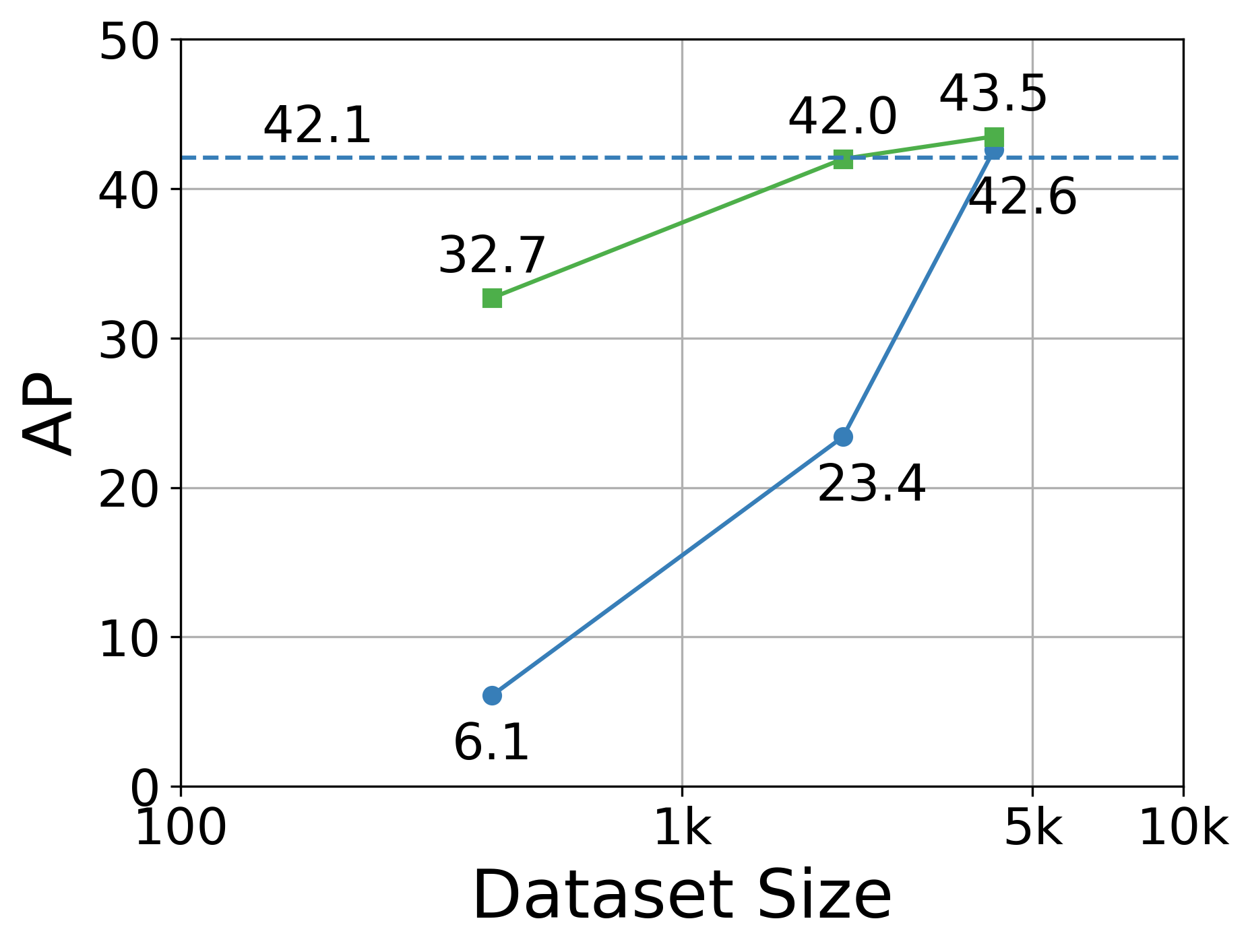}}
\subfigure[FAIR1M, IoU=0.25]{\label{fig:other25_student-c}\includegraphics[width=0.32\textwidth]{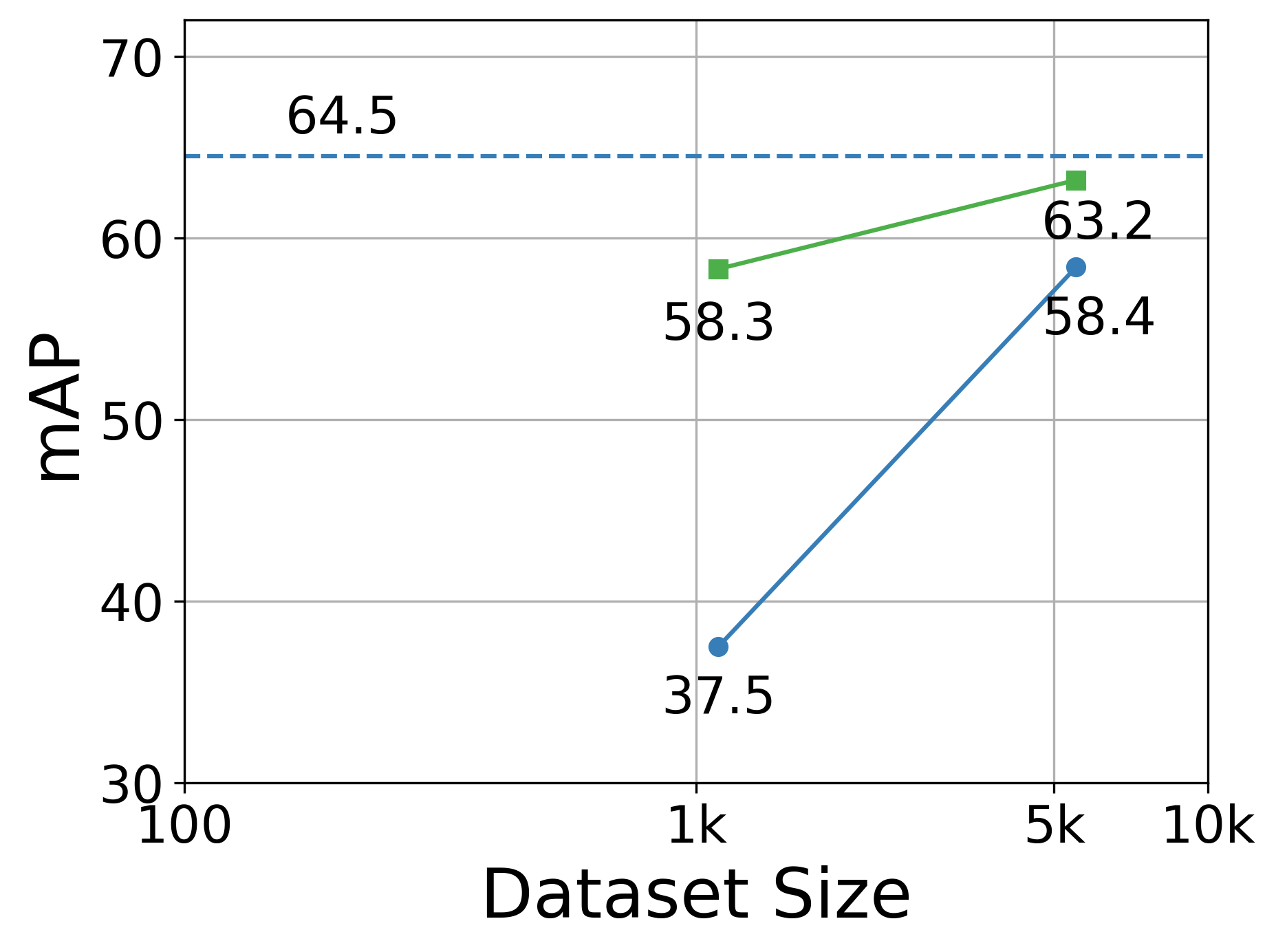}}
\subfigure[In-country Turbine, IoU=0.75]{\label{fig:other75_student-a}\includegraphics[width=0.32\textwidth]{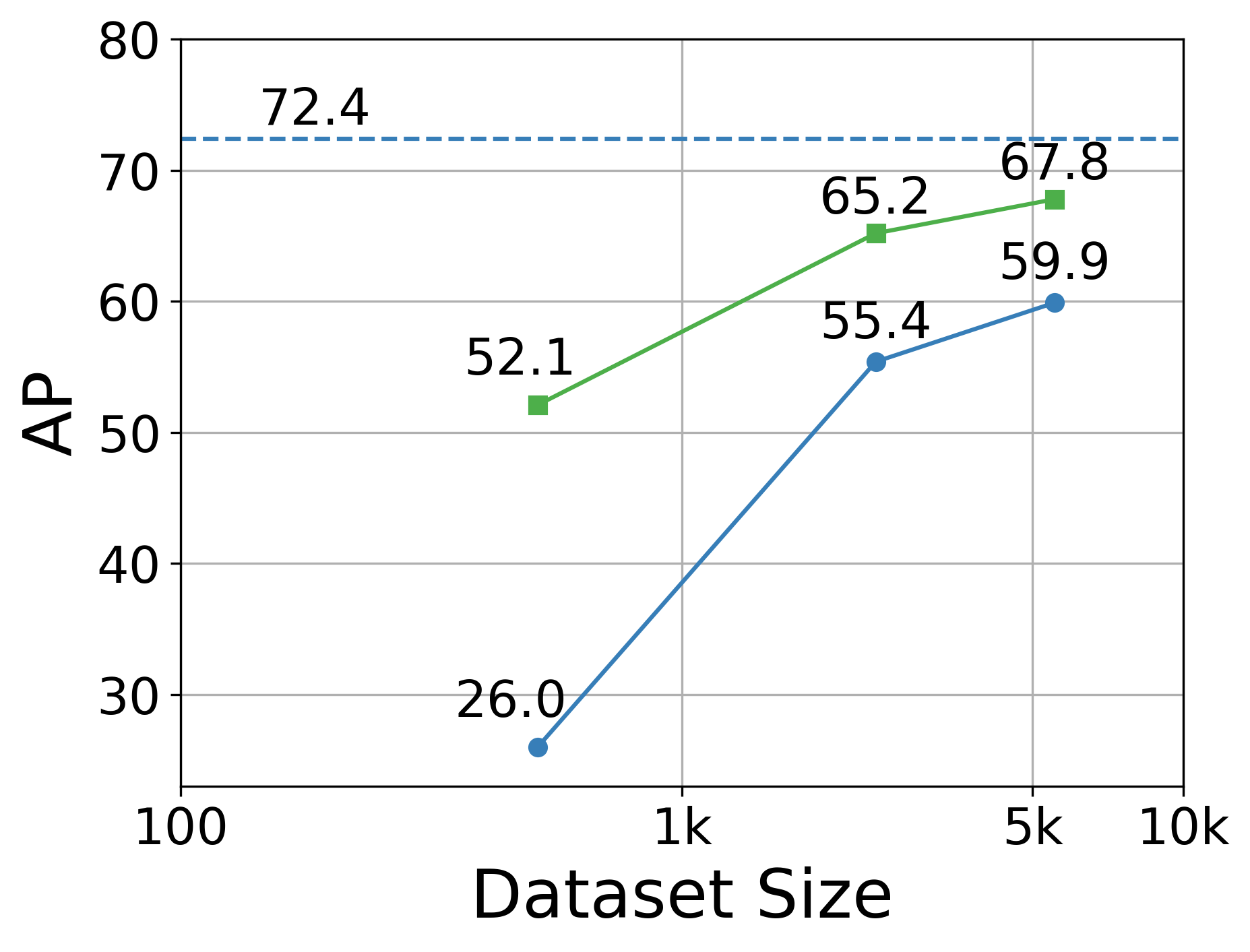}}
\subfigure[Out-country Turbine, IoU=0.75]{\label{fig:other75_student-b}\includegraphics[width=0.32\textwidth]{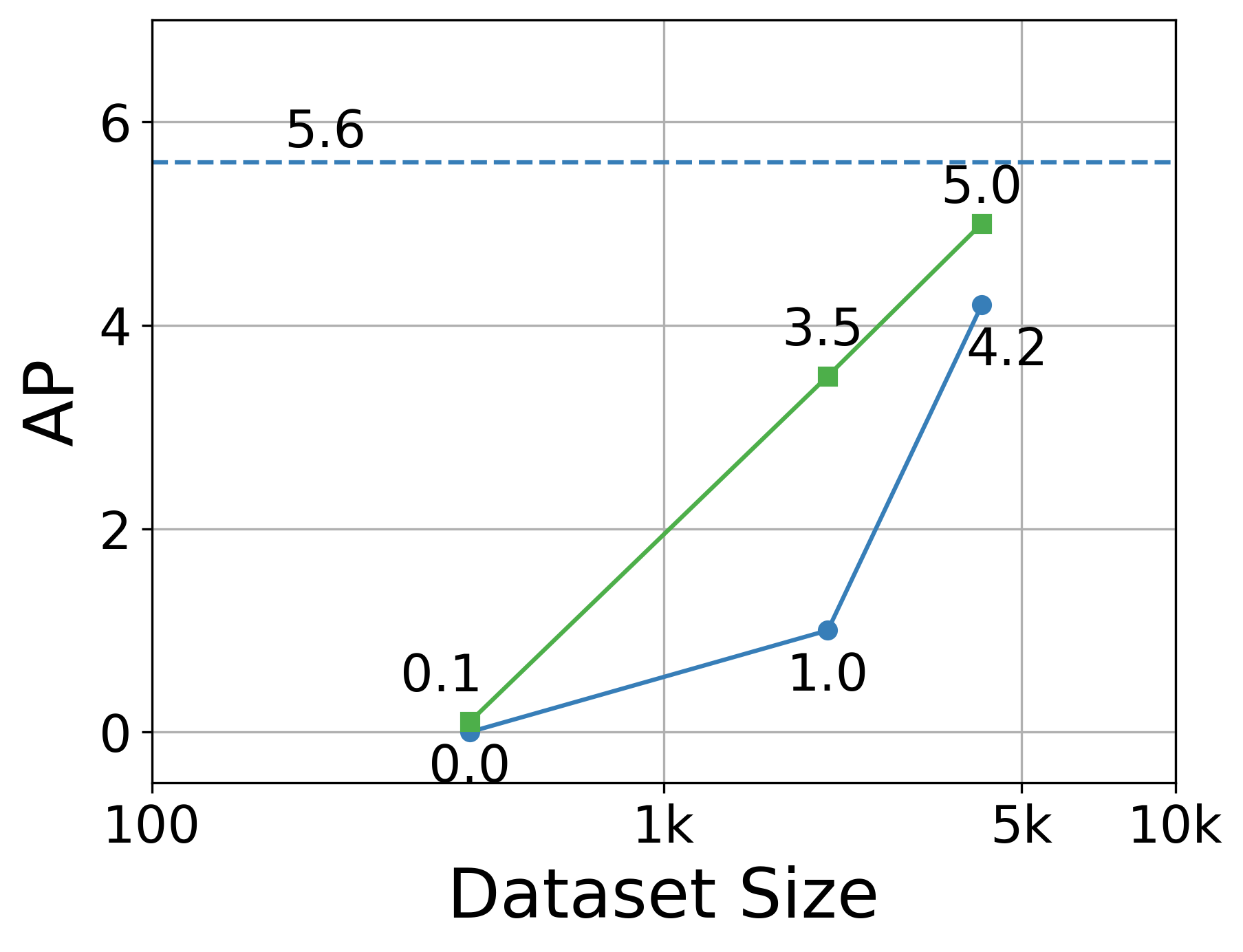}}
\subfigure[FAIR1M, IoU=0.75]{\label{fig:other75_student-c}\includegraphics[width=0.32\textwidth]{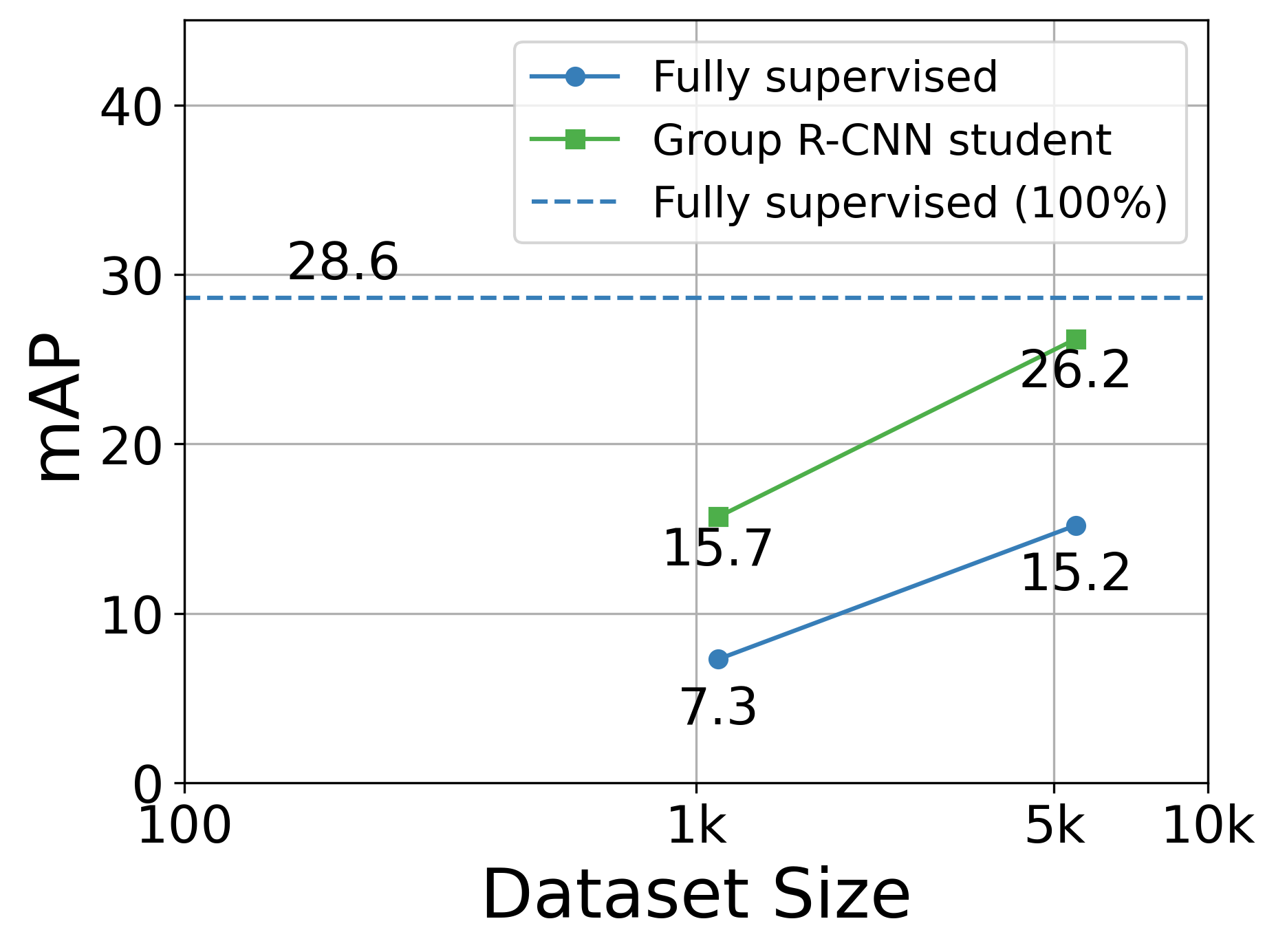}}
\caption{Comparison of Group R-CNN student model in comparison to fully supervised counterparts.}
\label{fig:other_student}
\end{figure*}

\end{document}